\documentclass[letterpaper, 10 pt, conference]{ieeeconf}  

\chardef\usethreeparttable=0
\chardef\usemathcommands=1
\chardef\usealgorithmicx=0
\chardef\usexcolor=0

\newcommand{\importifndef}[2]{\ifcsname#1\endcsname\else\usepackage{#2}\fi}
\newcommand{\tryusepackage}[1]{\IfPackageLoadedTF{#1}{}{\usepackage{#1}}}

\if\the\usexcolor
\usepackage[dvipsnames, svgnames, x11names, table]{xcolor}
\fi

\makeatletter
\importifndef{NAT@parse}{natbib}
\makeatother

\importifndef{algorithm}{algorithm}
\usepackage[noend]{algpseudocode}

\if\the\usealgorithmicx1
\tryusepackage{algorithmicx}
\fi

\tryusepackage{amsmath}
\tryusepackage{amsfonts}
\importifndef{proof}{amsthm}
\tryusepackage{bm}
\tryusepackage{nicefrac}

\tryusepackage{microtype}
\tryusepackage{wrapfig}
\tryusepackage{graphicx}
\tryusepackage{subcaption}

\importifndef{strip}{cuted}

\tryusepackage{array}
\tryusepackage{diagbox}
\tryusepackage{multirow}
\tryusepackage{booktabs}

\if\the\usethreeparttable1
\tryusepackage{threeparttable}
\fi

\usepackage{xspace}
\usepackage[acronym]{glossaries}
\makeglossaries

\usepackage{rotating}

\tryusepackage{tikz}
\tryusepackage{pgfplots}
\usetikzlibrary{plotmarks}
\pgfplotsset{compat=1.16}

\definecolor{wong1}{rgb}{0.9019607843137255, 0.6235294117647059, 0.0}
\definecolor{wong2}{rgb}{0.33725490196078434, 0.7058823529411765, 0.9137254901960784}
\definecolor{wong3}{rgb}{0.0, 0.6196078431372549, 0.45098039215686275}
\definecolor{wong4}{rgb}{0.9411764705882353, 0.8941176470588236, 0.25882352941176473}
\definecolor{wong5}{rgb}{0.0, 0.4470588235294118, 0.6980392156862745}
\definecolor{wong6}{rgb}{0.8352941176470589, 0.3686274509803922, 0.0}
\definecolor{wong7}{rgb}{0.8, 0.4745098039215686, 0.6549019607843137}

\pgfplotscreateplotcyclelist{cvl}{
solid, wong1, every mark/.append style={solid,fill=\pgfplotsmarklistfill}, mark=*\\
solid, wong2, every mark/.append style={solid,fill=\pgfplotsmarklistfill}, mark=square*\\
solid, wong3, every mark/.append style={solid,fill=\pgfplotsmarklistfill}, mark=triangle*\\
}

\tryusepackage{listings}

\definecolor{codegreen}{rgb}{0,0.6,0}
\definecolor{codegray}{rgb}{0.5,0.5,0.5}
\definecolor{codepurple}{rgb}{0.58,0,0.82}
\definecolor{backcolour}{rgb}{0.95,0.95,0.92}
\definecolor{dkgreen}{rgb}{0,0.6,0}
\definecolor{gray}{rgb}{0.5,0.5,0.5}
\definecolor{mauve}{rgb}{0.58,0,0.82}
 
\lstdefinestyle{lstStyleCode}{
    backgroundcolor=\color{backcolour},
    basicstyle=\ttfamily\footnotesize,
    breakatwhitespace=false,
    captionpos=b,
    keepspaces=true,
    numbers=left,
    numbersep=5pt,
    frame=single,
    showspaces=false,
    showstringspaces=false,
    showtabs=false,
    tabsize=2,
    xleftmargin=2em,
    xrightmargin=2em,
    breaklines=true,
    breakindent=2em,
    framexleftmargin=2em,
    framexrightmargin=2em,
    aboveskip=1em,
    columns=flexible,
    numberstyle=\tiny\color{gray},
    keywordstyle=\color{blue},
    commentstyle=\color{dkgreen},
    stringstyle=\color{mauve},
    breaklines=true,
}

\lstnewenvironment{codeblock}[1][]
    {\lstset{style=lstStyleCode,#1}}{}

\makeatletter
\let\NAT@parse\undefined
\makeatother
\tryusepackage{hyperref}
\importifndef{qty}{siunitx}

\importifndef{tablefootnote}{tablefootnote}
\usepackage{colortbl}

\let\labelindent\undefined
\importifndef{labelindent}{enumitem}
\tryusepackage{makecell}

\renewcommand{\sectionautorefname}{Sec.}

\newcommand{\bpm}[2]{\textbf{#1} \ensuremath{\pm} \textbf{#2}}
\newcommand{\npm}[2]{#1 \ensuremath{\pm} #2}

\newcommand{\fixed}[1]{\iffalse fixed: #1\fi}
\newcommand{\rej}[1]{\iffalse#1\fi}

\newcommand{\car}[2]{#1}

\newcommand{\minisection}[1]{\noindent\textbf{#1.~}}

\if\the\usemathcommands1

\def\eqref#1{Eq.~(\ref{#1})}

\def\1{\bm{1}}

\def\vb{{\bm{b}}}

\def\ve{{\bm{e}}}

\def\vq{{\bm{q}}}

\def\vs{{\bm{s}}}

\def\mM{{\bm{M}}}

\DeclareMathAlphabet{\mathsfit}{\encodingdefault}{\sfdefault}{m}{sl}
\SetMathAlphabet{\mathsfit}{bold}{\encodingdefault}{\sfdefault}{bx}{n}

\def\gA{{\mathcal{A}}}

\def\gC{{\mathcal{C}}}

\def\gE{{\mathcal{E}}}

\def\gO{{\mathcal{O}}}

\def\gQ{{\mathcal{Q}}}

\def\gS{{\mathcal{S}}}

\newcommand{\Ls}{\mathcal{L}}

\fi

\IEEEoverridecommandlockouts                              

\def\ours{H-Zero\xspace}

\title{\LARGE \bf
\ours: Cross-Humanoid Locomotion Pretraining\\Enables Few-shot Novel Embodiment Transfer
}

\ifx\peerreview\undefined
\author{
\authorblockN{
Yunfeng Lin$^{1,2}$,
Minghuan Liu$^{3,\dagger}$,
Yufei Xue$^{1}$,
Ming Zhou$^{2}$,
Yong Yu$^{1}$,
Jiangmiao Pang$^{2}$,
Weinan Zhang$^{1,2}$
}
\authorblockA{
\textsuperscript{1}Shanghai Jiao Tong University \quad \textsuperscript{2}Shanghai AI Lab \quad 
\textsuperscript{3}ByteDance Seed
}
\thanks{$^\dagger$Minghuan did the work when at ByteDance Seed.}
}
\else
\author{\authorblockN{Anonymous Authors}}
\fi

\begin{document}

\maketitle
\thispagestyle{empty}
\pagestyle{empty}


\begin{strip}
\begin{minipage}{\textwidth}\centering
\vspace{-20pt}
\includegraphics[width=1.0\textwidth]{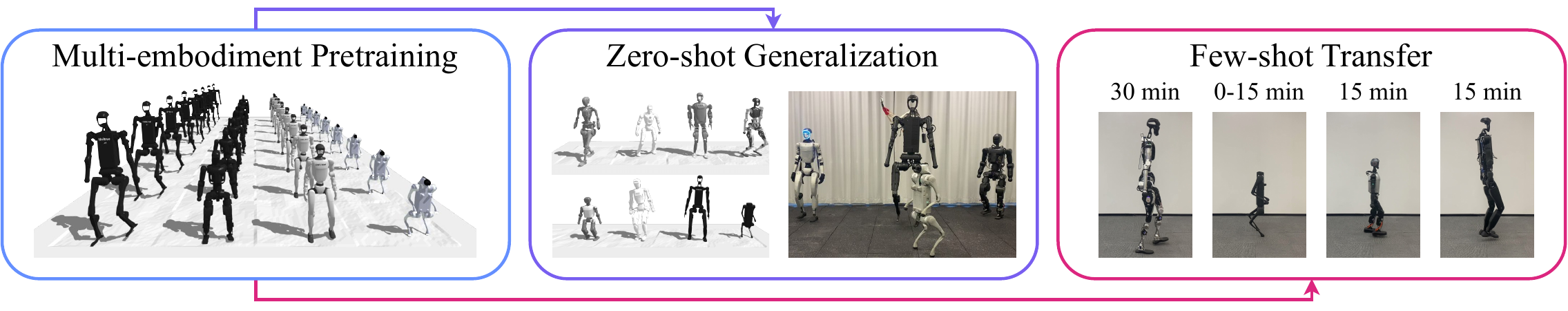}
\captionof{figure}{
\textbf{Left:} We propose a locomotion pretraining pipeline for humanoids by mixing multiple randomized embodiments into the training set.
\textbf{Middle:} The pretrained policy shows moderate adaptability to unseen embodiments and real hardware.
\textbf{Right:} Fine-tuning the pretrained policy achieves stable control on unseen robots with minimal additional training time.
}
\label{figurelabel}
\end{minipage}
\end{strip}

\begin{abstract}
The rapid advancement of humanoid robotics has intensified the need for robust and adaptable controllers to enable stable and efficient locomotion across diverse platforms.
However, developing such controllers remains a significant challenge because existing solutions are tailored to specific robot designs, requiring extensive tuning of reward functions, physical parameters, and training hyperparameters for each embodiment.
To address this challenge, we introduce \ours, a cross-humanoid locomotion pretraining pipeline that learns a generalizable humanoid base policy.
We show that pretraining on a limited set of embodiments enables zero-shot and few-shot transfer to novel humanoid robots with minimal fine-tuning.
Evaluations show that the pretrained policy maintains up to 81\% of the full episode duration on unseen robots in simulation while enabling few-shot transfer to unseen humanoids and upright quadrupeds within 30 minutes of fine-tuning.
\end{abstract}

\section{Introduction}

The rapid proliferation of novel and customized humanoid robot designs has intensified the demand for control algorithms capable of enabling robust and versatile locomotion across diverse embodiments. Humanoid robots, characterized by their anthropomorphic structure and high degrees of freedom (DoFs), offer unparalleled versatility in tasks such as bipedal walking, dynamic balancing, and complex manipulation. However, this anthropomorphic design also introduces significant challenges in controller development because of the intricate interplay of high-dimensional joint configurations, variable morphologies, and dynamic physical interactions.

Deep reinforcement learning (DRL) methods have shown promise in training locomotion controllers by leveraging physical simulations and trajectory sampling.
Combined with techniques such as domain randomization~\cite{peng2018}, knowledge distillation~\cite{rma, miki2022learning}, explicit models~\cite{fu2022deep, jenelten2023}, and curriculum learning~\cite{amp-wu}, policies trained in simulation can consistently transfer to the real world while maintaining performance\car{, efficiency~\cite{minimizing} and safety~\cite{he2024agile}}{}.
However, most existing approaches treat each robot variant as a distinct learning problem, resulting in policies tightly coupled to specific morphologies. Even minor variations in joint layouts, mass distributions, or limb dimensions often degrade performance, requiring resource-intensive retraining from scratch. As the diversity of humanoid platforms continues to grow, there is an urgent need for generalizable controllers that can adapt to new embodiments with minimal effort.

To address this challenge, we propose \ours, a novel pretraining pipeline for developing generalized locomotion policies for humanoid robots. Our approach introduces transformation layers at the policy's input and output to standardize control semantics across diverse humanoid embodiments, enabling unified policy representations. We further incorporate cross-embodiment diversity through randomized physical parameters, varied policy observations, diverse environment rollouts, and exploratory learning strategies. By integrating these techniques, \ours learns a robust base policy capable of controlling multiple humanoid models simultaneously. This pretrained policy can serve as a foundation for few-shot adaptation, allowing rapid fine-tuning to novel hardware with minimal data and computational resources.


We evaluate \ours on a comprehensive suite of humanoid robots with varying kinematic structures and physical properties, including simulated and real-world platforms. Our experiments demonstrate that the pretrained policy achieves superior few-shot adaptation, matching the performance of policies trained from scratch in only hundreds of epochs across benchmark locomotion tasks. Moreover, both the pretrained and fine-tuned policies exhibit consistent sim-to-real transfer, enabling robust performance on physical robots and highlighting the feasibility of shared policy learning for scalable humanoid control.

In summary, the contributions of this paper are:
\begin{itemize}
    \item A unified control interface that employs state transformations to standardize policy inputs and outputs across diverse humanoid embodiments.
    \item A cross-embodiment training environment with physical randomization and various training strategies to learn generalizable locomotion policies from a set of existing robots.
    \item Demonstrating that cross-embodiment locomotion pretraining enables zero-shot and few-shot adaptation to novel robots, significantly reducing the need for extensive retraining.
\end{itemize}


\section{Related Work}

\subsection{Legged locomotion via reinforcement learning}

Reinforcement learning provides a competent alternative for developing robot locomotion controllers by optimizing policies against task objectives instead of relying on manual implementation.
RL has been widely applied to various robots and tasks, including quadrupeds~\cite{lee2020learning, smith2022, rma, hoeller2024anymal}, wheeled robots~\cite{lee2024learning}, and humanoids~\cite{gu2024advancing,xue2025unified,wang2025beamdojo,huang2025learning}.
For quadrupeds, agile locomotion has been achieved over various terrains with both blind~\cite{tert,Daydreamer, margolis2023, caluwaerts2023} and exteroception-based RL policies~\cite{miki2022learning,parkour2}.
For humanoid robots, RL also enables whole-body control that achieves diverse gaits~\cite{he2024hover}, motion mimics~\cite{cheng2024express, ji2024exbody2,yin2025unitrackerlearninguniversalwholebody}, as well as integrations with upper-body teleoperation and manipulation~\cite{he2024learning, he2024omnih2o}.


\subsection{Cross-embodiment locomotion learning}

With the development of new robot models, recent research focuses on synthesizing controllers that generalize to hardware with varied shapes, weights, and even morphologies.
Some works propose to adopt specialized network architectures as a form of inductive bias.
For example, Graph Neural Networks (GNNs) can capture the robot's morphology by modeling a fixed set of joints as graph nodes~\cite{huang2020,whitman2023}.
On the other hand, transformer-based methods allow flexible observation and action dimensions by treating them as sequences~\cite{Yu2022MultiembodimentLR, bohlinger2024one,doshi2024scaling}.
This enables unified control across varied dimensions but demands high computational resources.

The combination of RL with other generative models such as diffusion~\cite{yang2025multiloco} has also been explored.
Beyond legged robots, cross-embodiment methods have been applied to dexterous robot hands~\cite{Patel2024GETZeroGE} to enable embodiment-aware manipulation.

Another line of work procedurally generates embodiments with randomized morphological and physical properties~\cite{feng2023genloco,ai2025embodimentscalinglawsrobot}, covering quadrupeds, hexapods, and humanoids.
This improves policy generalization as a form of domain randomization~\cite{imitate, randomization2, dreamwaq}.
However, determining the distribution of generated embodiments is crucial for policy performance and requires considerable expert knowledge for refinement.
As a result, real-robot applications remain challenging, especially for humanoids because of their control complexity.
In this work, we propose a locomotion pretraining framework for humanoids that enables rapid adaptation to real robots using less training time and fewer hyperparameters.

\newcommand{\es}{\mathbf{s}}
\newcommand{\qctrl}{\mathbf{\bar{q}}}
\newcommand{\q}{\mathbf{q}}
\newcommand{\qd}{\dot{\mathbf{q}}}
\newcommand{\qdd}{\ddot{\mathbf{q}}}
\newcommand{\x}{\mathbf{x}}
\newcommand{\xd}{\dot{\mathbf{x}}}
\newcommand\eM{M}
\newcommand\etau{{\mathbf{\tau}}}
\newcommand\eC{{\mathbf{C}}}
\newcommand\eJ{{\mathbf{J}}}
\newcommand\efC{{\mathbf{f_c}}}
\newcommand\etheta{{\mathbf{\theta}}}

\section{Preliminaries}


Humanoid locomotion control seeks to develop policies for stable and versatile movement in anthropomorphic robots, including walking, running, turning, and navigating complex terrain. This can be framed as a partially observable Markov decision process (POMDP) $M=(\gS, \gO, \gC, \gA, P, Z)$, where $\gS$, $\gO$, $\gC$, and $\gA$ denote the state, observation, command, and action spaces; $P$ is the transition function, and $Z$ the observation model. The goal is to learn a policy $\pi(o, c)$ that maps observations and commands to joint-level actions.



Reinforcement learning (RL) is commonly used to optimize such policies by maximizing expected cumulative reward. In whole-body humanoid control, the reward balances multiple objectives: tracking velocity commands (e.g., forward, lateral, and yaw), maintaining base height for balance, and aligning torso orientation. Additional terms penalize energy inefficiency, instability, or gait deviations to promote robust and efficient motion.

However, most RL-based controllers are morphology-specific, requiring tailored reward functions for each robot. Our framework addresses this by learning a generalizable base policy across embodiments, enabling few-shot adaptation to novel morphologies.

\section{Cross-Embodiment Humanoid Control}

\begin{figure*}[ht]
    \centering
    \includegraphics[width=0.95\linewidth]{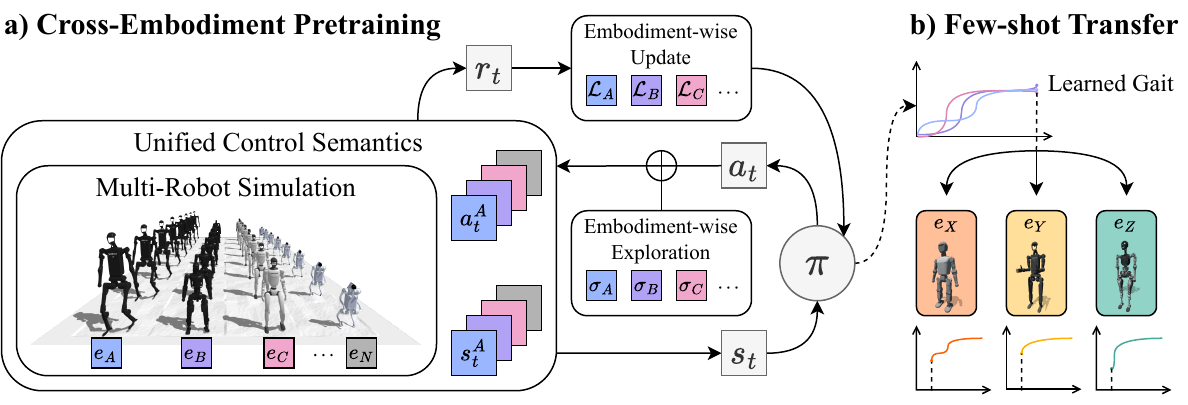}
    \caption{
\textbf{Method overview.}
\textbf{a)} The policy is pretrained by learning on a diverse set of humanoid embodiments through multi-robot simulation with unified control.
Training progress is dynamically balanced with embodiment-wise exploration and gradient updates.
\textbf{b)} At deployment, the pretrained policy supports few-shot adaptation to novel robots.
}
    \label{fig:overview}
\end{figure*}


To support generalization across diverse humanoid embodiments, we extend the POMDP formulation by augmenting the state space:
\begin{equation}
\gS = \gQ \times \gE~,
\end{equation}
where $\gQ$ captures the system's kinematic states (e.g., base position, joint angles, velocities), and $\gE$ represents embodiment-specific parameters such as joint layout, link dimensions, and inertial properties. During training, an embodiment $\ve \in \gE$ is sampled per rollout and held fixed, exposing the policy to a range of morphologies.

The observation space $\gO$ similarly includes both kinematic state and embodiment. The objective is to learn a unified policy that generalizes across embodiments by capturing shared control strategies.

\subsection{Unified Control Semantics}
\label{sec:ucs}

Learning policies for individual humanoid robots often relies on embodiment-specific joint configurations and control interfaces, resulting in incompatible action and observation spaces and poor transferability. To address this, we introduce a unified control representation that standardizes joint semantics across robots, enabling consistent policy learning and deployment.

We define a hardware-agnostic joint state space comprising current positions, velocities, and target positions for key rotational joints (e.g., head, shoulders, knees, ankles). Each robot maps its physical joints to this space based on kinematic roles—for example, identifying a wrist pitch joint as one rotating the hand link around the Y-axis from a reference pose.

This mapping is formalized as a bidirectional transformation between the robot's physical joint space and the unified environment space:

\begin{equation}
\vq_{\text{env}} = \mM \cdot \vq_{\text{phy}}, \quad \vq_{\text{phy}} = \mM^T \cdot \vq_{\text{env}}~,
\end{equation}

where \(\mM\)
encodes index assignments and zero-padding to accommodate varying degrees of freedom (DoFs)~\cite{yu2020,bohlinger2024onepolicy, yang2025multiloco}.

To ensure consistent motion across embodiments, we further apply kinematic alignment:

\begin{equation}
\vq_{\text{env}} = \mM \cdot (\vs \odot (\vq_{\text{phy}} - \vb))~,
\end{equation}

where 
\(\vs\)
 adjusts joint directions, and 
\(\vb\)
 offsets neutral positions to a standardized upright pose. This alignment ensures that joint commands produce equivalent physical motions across diverse morphologies, supporting coherent locomotion behaviors in both simulation and real-world deployment.

\subsection{Embodiment Descriptors}
\label{sec:desc}

The unified control semantics introduced above enable cross-embodiment control by standardizing the input and output spaces of a single policy with fixed parameter dimensions. Typical simulation environments, however, are designed to train standalone policies tailored to individual robot variants, making them inherently hardware-specific. In such settings, controllers rely solely on proprioceptive observations—such as joint positions, velocities, and inertial measurement unit (IMU) readings—without explicit awareness of the robot's configuration because these physical properties are assumed to be static and implicit in the training process.

In contrast, a cross-embodiment policy must dynamically adapt to the physical properties of the controlled robot to generate optimal actions for stable and efficient locomotion. To achieve embodiment-aware control, we introduce embodiment descriptors: compact, informative features that encode key aspects of each robot's physical properties.
These descriptors span multiple domains, including kinematics (\textit{e.g.}, joint limits, rotation axes), topology (\textit{e.g.}, hierarchical joint structures), geometry (\textit{e.g.}, link lengths, shapes), and dynamics (\textit{e.g.}, mass, inertia, stiffness, damping).
The resulting joint and rigid body features are arranged according to the unified space in~\sectionautorefname~\ref{sec:ucs} to ensure consistent semantics across robots.
\begin{equation}
\begin{aligned}
Z_{\text{JD}}(\ve) &= \mM\cdot[(K_p,K_d,\tau_{max})_{1\dots n}]^T\\
Z_{\text{RD}}(\ve) &= \mM\cdot[(m,h,I)_{1\dots n}]^T\\
Z_\text{e}(\ve) &= [Z_{\text{JD}}(\ve), Z_{\text{RD}}(\ve), Z_{\text{Kine}}(\ve), Z_{\text{Geom}}(\ve)]
\end{aligned}
\end{equation}




These additional states beyond partial observations can either be treated as privileged information, accessible only to the value function in asymmetric actor-critic architectures~\cite{choi2023}, or be incorporated directly into the policy observation space because the descriptors are available at inference time.
We compare these setups in the evaluation section.

\subsection{Embodiment Training Set}
\label{sec:mix}

To train a cross-embodiment policy capable of controlling multiple humanoid robots, we generate simulated trajectories using a diverse set of robot models, leveraging unified control semantics and embodiment descriptors. The policy is optimized to perform locomotion tasks across all embodiments concurrently.

While prior work applies domain randomization (DR) to perturb system parameters such as joint friction and mass~\cite{Valassakis2020CrossingTG,campanaro2022,8202133}, we extend this technique by broadening the DR range to better approximate the dynamics of varied robot morphologies. Specifically, we double the perturbation bounds for rigid-body and joint dynamics, while structural diversity (e.g., geometry and topology) is introduced by mixing existing robot models.

Unlike approaches that procedurally generate randomized hardware from scratch~\cite{ai2025embodimentscalinglawsrobot}, which often require extensive tuning of physical hyperparameters, our method relies on targeted extrapolation from real-world designs. This yields strong generalization performance with minimal engineering overhead, as demonstrated in our experiments.

\subsection{Embodiment-Aware Learning}
\label{sec:var}

We adopt a unified reward formulation across all robots, following the design in~\cite{xue2025unified}. While the structure of the reward function remains identical, its coefficients—such as nominal base height and stance width—are tailored to each embodiment.
This multi-robot setup presents greater learning challenges than single-robot training due to diverse dynamics, stability profiles, and task complexities.

To address this, we introduce an embodiment-aware training strategy that assigns independent exploration sampling variances \(\sigma\) to each robot type.
These variances are updated adaptively based on gradients from each embodiment, enabling tailored exploration: robots exhibiting slower learning progress receive higher variance to promote broader action sampling, while more stable embodiments converge under reduced variance.

In addition to adaptive exploration, we dynamically reweight the loss of each embodiment type based on their current performance, quantified by average episodic return.

\begin{equation}
    \mathcal{L}_\text{total}=\sum_{i=1}^N w_i \Ls_i = \sum_{i=1}^N (1 - \frac{R_i - R_\text{min}}{R_\text{max} - R_\text{min} + \epsilon}) \Ls_i
\end{equation}


This dynamic loss scheduling ensures that underperforming embodiments receive increased gradient emphasis during training.
Together, these techniques promote balanced learning progress across the embodiment set, accelerating overall convergence and enhancing policy robustness across a wide range of humanoids and even standing quadrupeds.






\subsection{Cross-Embodiment Transfer}





Leveraging a specialized learning environment and task-aware training algorithm, \ours learns a family of humanoid locomotion policies that generalize across robot variants, serving as a pretrained controller for rapid transfer.

The framework begins by curating a diverse set of robot embodiments with varying structures (e.g., joint layouts, limb proportions) and dynamics (e.g., mass, inertia). During pretraining, the policy learns to control these embodiments using extended domain randomization, unified semantics, and embodiment descriptors, yielding a robust base policy.

To transfer to a novel robot, we perform few-shot fine-tuning using a small set of trajectories.
A key design is to use a low action variance during initial sampling, as excessive noise can destabilize walking behaviors and hinder adaptation.
This approach supports zero-shot inference and few-shot adaptation on similar morphologies, and enables sim-to-real deployment. In experiments, \ours outperforms baselines trained from scratch in both task performance and learning efficiency on unseen robots.

\section{Evaluations}

\begin{table}[t]

\centering
\caption{Robot models used in our experiments}
\label{tab:hardware}
\begin{tabular}{lc@{\enspace}c @{\enspace}c @{\enspace}ccc} 
\toprule
\multirow{2}{*}{\textbf{Model Name}} & \multirow{2}{*}{\textbf{Alias}} & \multicolumn{3}{c}{\textbf{DoFs}} & \multicolumn{1}{l}{\multirow{2}{*}{\textbf{Mass}}} & \multirow{2}{*}{\begin{tabular}[c]{@{}c@{}}\textbf{Base}\\\textbf{Height}\end{tabular}} \\ 
\cmidrule{3-5}
 &  & \textbf{Arm} & \textbf{Waist} & \textbf{Leg} & \multicolumn{1}{l}{} &  \\ 
\midrule
Unitree~\cite{unitree}~H1 Gen1 & H1 & 4 & 1 & 5 & 51.6 & 0.98 \\
Unitree G1 EDU & G1 & 7 & 3 & 6 & 33.3 & 0.78 \\
PND~\cite{pnd} Adam Lite & Adam & 5 & 3 & 6 & 58.4 & 0.88 \\
Fourier~\cite{fftai} GR-1 Pro & GR1 & 7 & 3 & 6 & 56.9 & 0.91 \\
Fourier N1 & N1 & 5 & 1 & 6 & 39.7 & 0.68 \\
Booster T1~\cite{t1} & T1 & 4 & 1 & 6 & 31.6 & 0.65 \\
EngineAI PM01~\cite{pm01} & PM01 & 5 & 1 & 6 & 40.9 & 0.82 \\
OpenLoong~\cite{oghr2} & OGHR & 4 & 1 & 6 & 75.9 & 1.12 \\
Leju Kuavo S42~\cite{kuavo} & Kv & 7 & 0 & 6 & 56.9 & 0.85 \\
Dobot Atom~\cite{atom} & Atom & 2 & 0 & 6 & 58.8 & 0.88 \\
Unitree H1 Gen2 & H1-2 & 4 & 1 & 5 & 67.3 & 0.96 \\ 
\midrule
Unitree Go2 & Go2 & 3 & 0 & 3 & 15.0 & 0.54 \\
Unitree A1 & A1 & 3 & 0 & 3 & 12.4 & 0.48 \\
Unitree Aliengo & AGo & 3 & 0 & 3 & 24.9 & 0.60 \\
\bottomrule
\end{tabular}

\end{table}

\begin{figure*}[htp]
\centering
\vspace{-5pt}
\begin{minipage}[t]{\textwidth}
    \centering
    \includegraphics[width=1\textwidth]{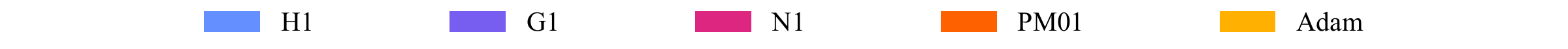}
\end{minipage}
\vspace{5pt}
    \begin{minipage}[t]{0.64\textwidth}
    \begin{minipage}[t]{0.5\textwidth}
        \centering
        \includegraphics[width=1\textwidth]{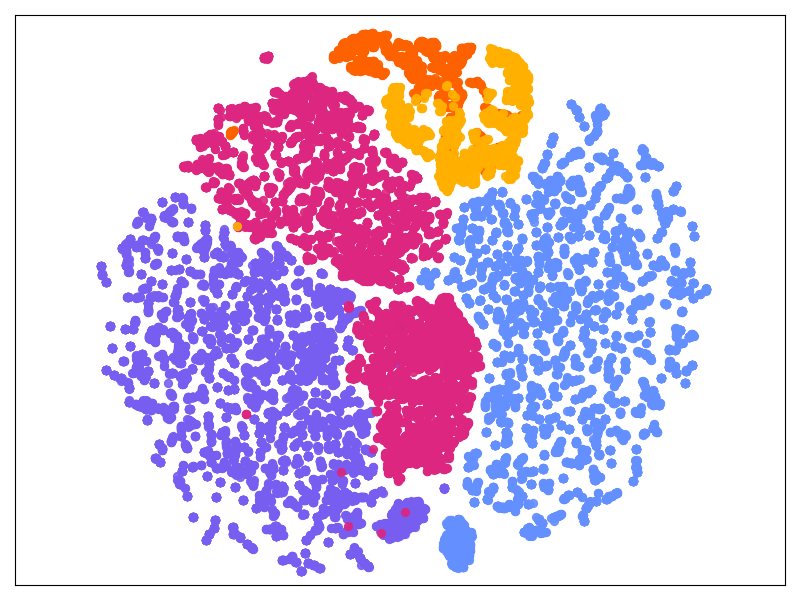}
        \vspace{-15pt}
    \end{minipage}
    \begin{minipage}[t]{0.5\textwidth}
        \includegraphics[width=1\textwidth]{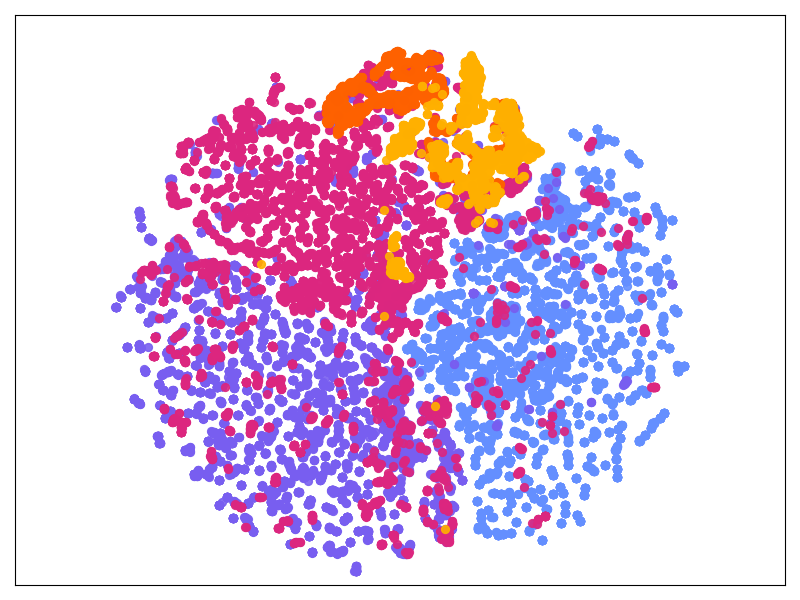}
        \vspace{-15pt}
    \end{minipage}
\caption{
\textbf{t-SNE~\cite{vandermaaten08a} visualization of rollout trajectories and embodiment descriptors under different domain randomization (DR).}
\textbf{Left}: standard DR for single-robot training.
\textbf{Right}: extended DR proposed in~\sectionautorefname~\ref{sec:mix} in cross-embodiment pretraining.
With extended DR, trajectories from unseen robots without randomization overlap with the broadened training distribution, demonstrating improved transferability.
}
    \label{fig:tsne_traj}
    \end{minipage}
    \hspace{4pt}
    \begin{minipage}[t]{0.32\textwidth}
    \begin{minipage}[t]{1\textwidth}
        \includegraphics[width=1\textwidth]{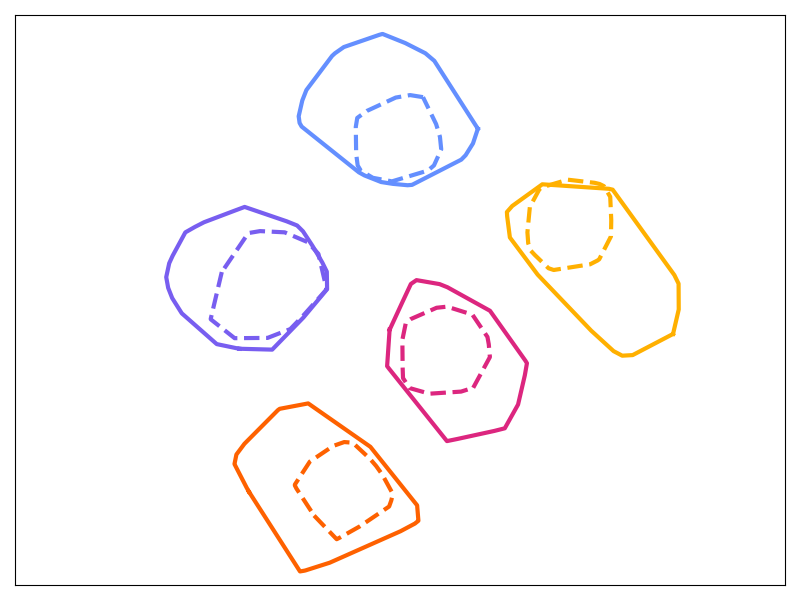}
        \vspace{-15pt}
    \end{minipage}
\caption{
t-SNE convex hulls of embodiment parameters under standard (dashed) and quadrupled (solid) DR ranges, which retain clear boundaries even under strong randomization.
}
    \label{fig:tsne_embd}
    \end{minipage}
\end{figure*}

\begin{table*}

\centering
\renewcommand{\arraystretch}{0.9}
\setlength{\extrarowheight}{0pt}
\addtolength{\extrarowheight}{\aboverulesep}
\addtolength{\extrarowheight}{\belowrulesep}
\setlength{\aboverulesep}{0pt}
\setlength{\belowrulesep}{0pt}

\caption{Cross-embodiment generalization performance of pretrained policies}
\label{tab:pretrain}
\begin{tabular}{lccccccccccccccc} 
\toprule
\multicolumn{1}{c}{\multirow{2}{*}{\textbf{Settings}}} & \multicolumn{14}{c}{\textbf{Evaluation Robot }} & \multirow{2}{*}{\begin{tabular}[c]{@{}c@{}}\textbf{Test}\\\textbf{Mean}\end{tabular}} \\ 
\cmidrule{2-15}
\multicolumn{1}{c}{} & H1 & N1 & G1 & GR1 & PM01 & Kv & Adam & T1 & OGHR & Atom & H1-2 & Go2 & A1 & \multicolumn{1}{l}{AGo} &  \\ 
\midrule
\multicolumn{16}{l}{{\cellcolor[rgb]{0.925,0.925,0.925}}\textbf{Ablation A: Training Set }} \\
H1 & 1.00 & 0.07 & 0.08 & 0.26 & 0.26 & 0.21 & 0.07 & 0.07 & 0.28 & 0.04 & 0.99 & 0.10 & 0.12 & \multicolumn{1}{l}{0.03} & 0.20 \\
H1,N1 & 1.00 & 1.00 & 1.00 & \textbf{0.67} & 0.87 & 0.90 & 0.12 & \textbf{0.99} & 0.41 & 1.00 & 0.99 & 0.12 & 0.12 & 0.03 & 0.60 \\
H1,N1,GR1 & 1.00 & 1.00 & 1.00 & 0.84 & 0.82 & 0.83 & 0.78 & 0.54 & \textbf{0.73} & 1.00 & 0.99 & 0.11 & 0.13 & 0.03 & 0.63 \\
H1,N1,G1 & 1.00 & 0.99 & 1.00 & 0.58 & \textbf{0.99} & 0.54 & \textbf{0.85} & 0.48 & 0.71 & \textbf{1.00} & 0.87 & 0.15 & 0.14 & 0.03 & 0.58 \\
H1,N1,G1,A1 & 1.00 & 0.99 & 0.99 & 0.48 & 0.91 & 0.86 & 0.17 & \textbf{0.99} & 0.71 & 0.97 & \textbf{1.00} & \textbf{0.99} & 0.99 & 0.70 & \textbf{0.81} \\
H1,N1,G1,GR1,Go2 & 1.00 & 1.00 & 1.00 & 1.00 & 0.58 & \textbf{1.00} & 0.36 & 0.72 & 0.64 & 0.85 & 0.95 & 0.95 & \textbf{0.99} & \textbf{0.99} & 0.79 \\
\multicolumn{16}{l}{{\cellcolor[rgb]{0.925,0.925,0.925}}\textbf{Ablation B: Domain Randomization }} \\
Single-robot range & 1.00 & 1.00 & 1.00 & 0.29 & 0.26 & 0.52 & 0.76 & 0.21 & 0.59 & 0.99 & 0.55 & 0.13 & 0.13 & 0.03 & 0.53 \\
Quadrupled range & 1.00 & 1.00 & 1.00 & 0.47 & 0.16 & 0.82 & 0.11 & \textbf{1.00} & 0.61 & 0.85 & 0.92 & 0.11 & 0.10 & 0.07 & 0.47 \\
\multicolumn{16}{l}{{\cellcolor[rgb]{0.925,0.925,0.925}}\textbf{Ablation C: Embodiment Descriptors }} \\
No Descriptor & 1.00 & 1.00 & 1.00 & \textbf{0.68} & 0.25 & 0.70 & 0.09 & 0.38 & 0.23 & 0.99 & 0.98 & 0.14 & 0.14 & \multicolumn{1}{l}{0.03} & 0.54 \\
Observable Desc. & 1.00 & 1.00 & 1.00 & 0.25 & 0.29 & 0.10 & 0.07 & 0.62 & 0.18 & 0.08 & 0.99 & 0.08 & 0.09 & \multicolumn{1}{l}{0.02} & 0.41 \\
\multicolumn{16}{l}{{\cellcolor[rgb]{0.925,0.925,0.925}}\textbf{Ablation D: Action space size}} \\
Whole-body (32) & 1.00 & 1.00 & 1.00 & 0.24 & 0.51 & 0.44 & 0.73 & 0.11 & 0.47 & 0.70 & 0.76 & 0.09 & 0.09 & 0.08 & 0.56 \\
Legs + Waist (15) & 1.00 & 1.00 & 1.00 & 0.42 & \textbf{0.99} & 0.76 & 0.08 & 0.62 & 0.03 & \textbf{1.00} & 0.96 & 0.11 & 0.03 & 0.09 & 0.46 \\

\bottomrule
\end{tabular}

\end{table*}

We focus on three research questions through simulation and real-world evaluations:

\begin{enumerate}[leftmargin=14pt]

    \item How does the performance of the pretrained policy scale with the number and diversity of training embodiments?
    \item How efficiently does the pretrained policy adapt to novel embodiments in few-shot settings?
    \item How well does the policy generalize to real-world deployment across different hardware platforms?
\end{enumerate}

\minisection{Network architecture}
We employ a simple multilayer perceptron (MLP) with ELU~\cite{elu} activation for the actor and critic network, which takes as input a five-step history of observations.
A state estimator predicts the base linear velocity to improve command tracking~\cite{xiao2025learning}.

\minisection{Training setups and robots involved}
Training is conducted using the Isaac Gym~\cite{isaacgym} physics simulator, running on a single NVIDIA GeForce RTX 4090 GPU.
The x-y-yaw velocity commands are randomly sampled within the ranges: \(v_x\in[-0.6, 1.2], v_y\in[-0.4, 0.4], v_\psi\in[-1.0, 1.0]\), with a curriculum that begins at half these ranges to facilitate early learning.
The set of robot embodiments used for training and evaluation is summarized in~\tableautorefname~\ref{tab:hardware}.

\noindent\textbf{Metrics.}
To quantify locomotion performance across diverse humanoid embodiments, we use the following metrics to assess the policy's ability to maintain stable and accurate motion across different embodiments:
\begin{itemize}[leftmargin=*]
    \item Normalized episode length: the ratio of executed policy steps to the episode limit, indicating transferability and behavioral stability on novel robots.
    \item $E_{v_{x}}, E_{v_{y}}$: error in the robot's root linear velocity across the horizontal plane, measuring deviation from commanded forward and lateral velocities.
    \item $E_{v_\psi}$: error in the root angular velocity about the z-axis, evaluating the accuracy of turning motions.
\end{itemize}
Tracking errors are reported as mean values per episode, excluding the first 50 steps to avoid transient initialization effects.





\subsection{Cross-embodiment pretraining}
To assess the role of embodiment diversity in pretraining, we compare policies trained on different subsets of robot models.
Training is conducted across 8192 parallel environments, evenly distributed among the selected embodiments.
All policies are trained for 50,000 epochs using the same reward functions.
Unless otherwise noted, the default training set consists of H1, N1, and G1; the action space is limited to 12 leg joints; the randomization range is doubled relative to single-robot settings; and embodiment descriptors are provided only to the critic.
Evaluations are conducted at 60\% of the velocity command ranges across 2048 environments for four episodes, with domain randomization disabled to isolate effects from embodiment differences.
Results are summarized in~\tableautorefname~\ref{tab:pretrain} and analyzed below.

\begin{figure}
    \centering
    \includegraphics[width=1\linewidth]{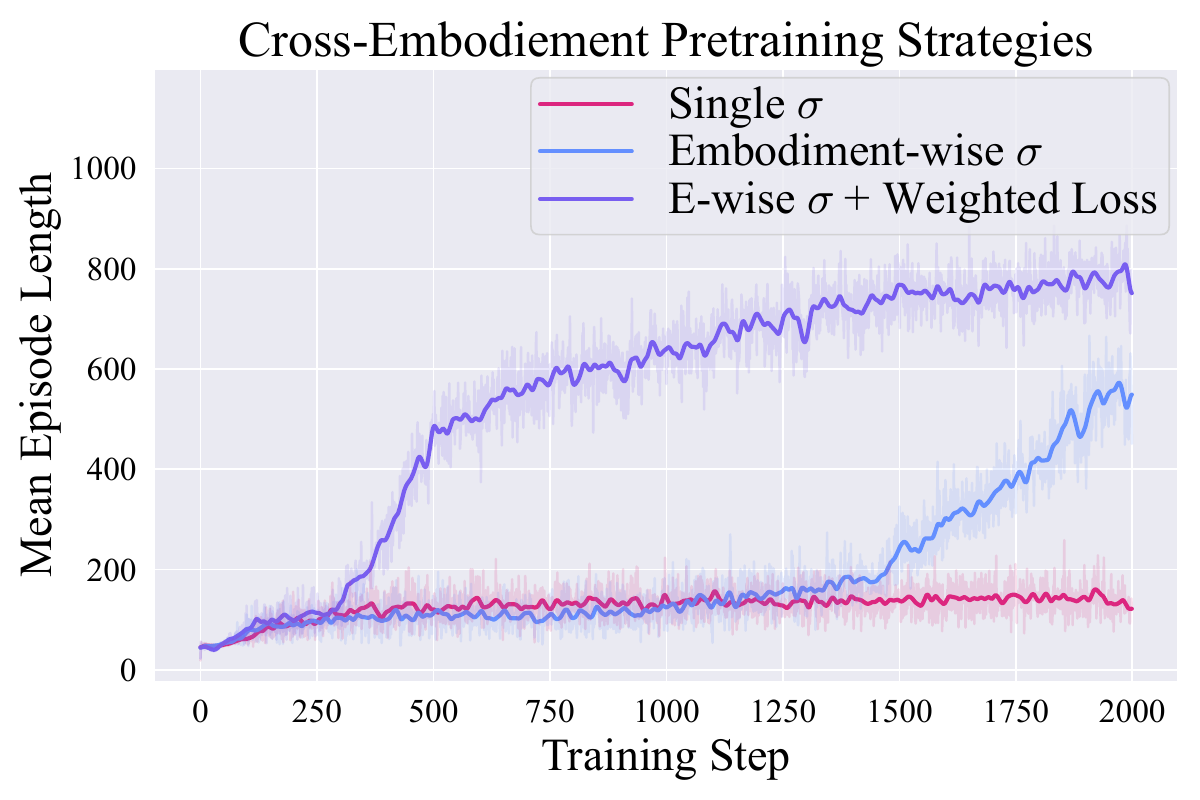}
    \caption{Mean episode length of cross-embodiment trainings (H1, N1, G1, and A1) under different training strategies.}
    \label{fig:var}
\end{figure}

\minisection{Embodiment training set}
We vary the composition of the training set to study the effect of embodiment mixing, as detailed in~\sectionautorefname~\ref{sec:mix}.
Results in \tableautorefname~\ref{tab:pretrain} show that policies trained on a broader set of embodiments consistently achieve higher performance on unseen robots than those trained on fewer embodiments, especially when the training set includes morphologically distinct robots.
We further validate the effectiveness of our training strategies in \sectionautorefname~\ref{sec:var}.
As visualized in \figureautorefname~\ref{fig:var}, dynamically adjusting sampling variances and loss scales across embodiments leads to more balanced learning progress.


\minisection{Domain randomization}
Ablation B confirms that expanding the domain randomization range improves transferability to unseen robots.
To visualize this effect, we collect the trajectories of pretrained policy rollouts under different randomization settings, and visualize them in~\figureautorefname~\ref{fig:tsne_traj}.
Each point represents a five-step state sequence sampled from a successful trajectory of one specific embodiment.
The first three robot models from the training set are randomized, whereas the last two unseen robots use no randomization.
Under single-embodiment randomization, trajectories from unseen robots show little similarity to the training set.
Extending the randomization range expands the coverage and begins to overlap with unseen robots, indicating better transferability across embodiments.

\minisection{Embodiment descriptor}
Ablation C shows that policies using observable embodiment descriptors perform worse than those relying on privileged descriptors.
To understand this, we visualize the extracted system parameters in~\figureautorefname~\ref{fig:tsne_embd}.
Despite randomization, the parameters form disjoint clusters per robot, which is expected since the embodiment geometry and structure are not randomized.
This also suggests that similar physical effects may arise from distinct configurations (e.g., higher torque compensating for increased mass).

\minisection{Action space size}
Expanding the action space to include waist and upper-body control has a negative impact on performance.
This is likely due to increased dimensionality and weak reward coupling for upper-body joints, signaling the need for more targeted objectives and regularization.

\subsection{Few-shot transfer}

\begin{table}
\centering
\caption{Per-embodiment tracking performance}
\label{tab:transfer}
\begin{tabular}{llcccc} 
\toprule
Robot & \multicolumn{1}{c}{\begin{tabular}[c]{@{}c@{}}Training\\Epochs\end{tabular}} & Return \(\uparrow\) & \begin{tabular}[c]{@{}c@{}}$E_{v_x}\downarrow$\\(cm/s)\end{tabular} & \begin{tabular}[c]{@{}c@{}}$E_{v_y}\downarrow$\\(cm/s)\end{tabular} & \begin{tabular}[c]{@{}c@{}}$E_{v_\phi}\downarrow$\\(deg/s)\end{tabular} \\ \midrule
\multirow{5}{*}{Adam} & S \quad 20k & \npm{53}{1.4} & \npm{4.6}{2.9} & \npm{6.9}{4.4} & \npm{6.0}{0.9} \\
 & S \quad 50k & \npm{52}{1.2} & \npm{3.9}{2.0} & \bpm{5.6}{2.1} & \bpm{3.1}{0.5} \\
 & P \quad 0 & \npm{20}{5.6} & \npm{22}{19} & \npm{13}{5.6} & \npm{13}{11} \\
 & P \quad 1k & \npm{49}{1.8} & \npm{4.8}{3.3} & \npm{8.1}{3.0} & \npm{4.3}{0.8} \\
 & P \quad 2k & \bpm{52}{1.5} & \bpm{3.7}{3.0} & \npm{7.2}{2.1} & \npm{3.5}{0.5} \\ 
\midrule
\multirow{5}{*}{{T1}} & S \quad 25k & \npm{48}{2.5} & \npm{9.0}{8.9} & \npm{10}{2.5} & \npm{7.6}{1.0} \\
 & S \quad 50k & \npm{49}{2.3} & \npm{8.8}{8.4} & \npm{12}{2.9} & \npm{7.8}{0.6} \\
 & P \quad 0 & \npm{37}{5.6} & \npm{20}{7.6} & \npm{16}{3.6} & \npm{27}{4.0} \\
 & P \quad 500 & \npm{50}{2.9} & \npm{13}{10} & \npm{10}{2.4} & \bpm{5.8}{0.4} \\
 & P \quad 2k & \bpm{52}{1.9} & \bpm{8.2}{8.0} & \npm{10}{2.7} & \npm{6.5}{0.4} \\ 
\midrule
\multirow{5}{*}{{H1-2}} & S \quad 25k & \npm{47}{6.2} & \npm{14}{17} & \npm{8.9}{4.1} & \npm{5.3}{1.8} \\
 & S \quad 50k & \npm{51}{1.0} & \npm{5.7}{1.7} & \bpm{7.8}{1.0} & \npm{4.7}{0.8} \\
 & P \quad 0 & \npm{31}{3.2} & \npm{17}{5.8} & \npm{10}{1.1} & \npm{8.0}{0.3} \\
 & P \quad 500 & \npm{45}{4.5} & \npm{5.2}{3.4} & \npm{9.7}{1.8} & \npm{5.5}{0.8} \\
 & P \quad 5k & \bpm{53}{1.1} & \bpm{4.3}{1.8} & \npm{8.4}{1.3} & \bpm{4.3}{0.4} \\ 
\midrule
\multirow{4}{*}{{AGo}} & S \quad 25k & \npm{50}{2.7} & \npm{18}{12} & \npm{12}{6.3} & \npm{18}{4.2} \\
 & S \quad 50k & \npm{53}{2.8} & \npm{17}{11} & \bpm{12}{6.2} & \npm{14}{3.8} \\
 & P \quad 0 & \npm{53}{3.0} & \bpm{11}{6.1} & \npm{13}{6.4} & \npm{13}{1.6} \\ 
 & P \quad 500 & \npm{53}{3.0} & \npm{13}{8.4} & \npm{12}{6.7} & \bpm{13}{1.8} \\ 
\bottomrule
\multicolumn{6}{l}{\makecell[l]{\footnotesize \textit{Notation:} S = Scratch initialization; P = Pretrained initialization; \\Epochs indicate retraining/fine-tuning duration.}}
\end{tabular}
\end{table}

We assess the transferability of pretrained policies to unseen robot models.
Specifically, we compare the adaptation performance of policies initialized from pretrained weights versus those trained from scratch.
Thanks to the unified action and observation spaces, weight reuse is straightforward.
Transfer performance is evaluated based on cumulative reward and tracking error after a fixed number of training epochs, as shown in~\tableautorefname~\ref{tab:transfer}.
Training configurations are denoted as a single letter followed by the number of training epochs onward: S indicates training from scratch, while P denotes fine-tuning from a pretrained policy.
For example, P 0 refers to direct evaluation of the pretrained policy without additional training.



\begin{figure}
    \centering
    \includegraphics[width=1\linewidth]{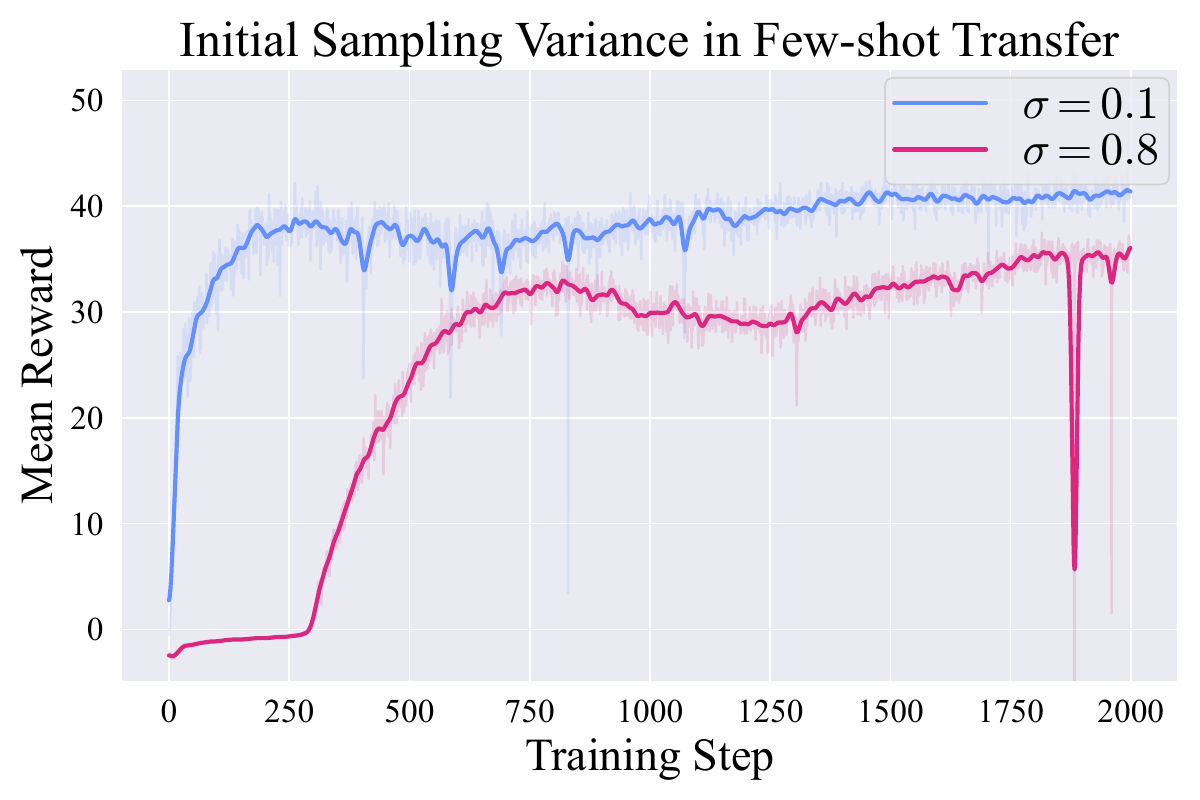}
    \caption{Few-shot transfer reward curves of PND Adam robot under different initial action sampling variance.}
    \label{fig:log}
\end{figure}

\def\figwidth{0.23}

\begin{figure*}[t]
    \centering
    \setlength{\tabcolsep}{2pt}
    \renewcommand{\arraystretch}{1.2}
    \begin{tabular}{c c c c c}
        & \textbf{Aliengo} & \textbf{T1} & \textbf{H1-2} & \textbf{Adam Lite} \\
        {\centering\begin{sideways}\textbf{Scratch}\end{sideways}}
        &
        \includegraphics[width=\figwidth\textwidth]{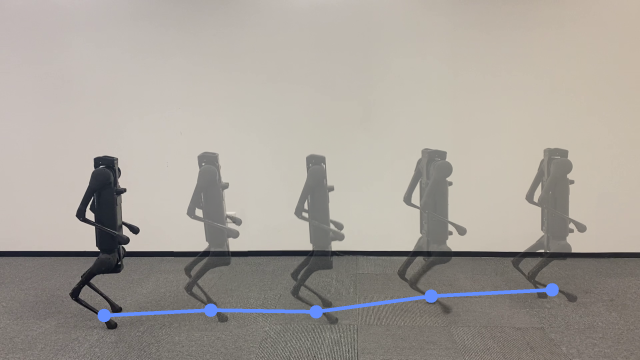} &
        \includegraphics[width=\figwidth\textwidth]{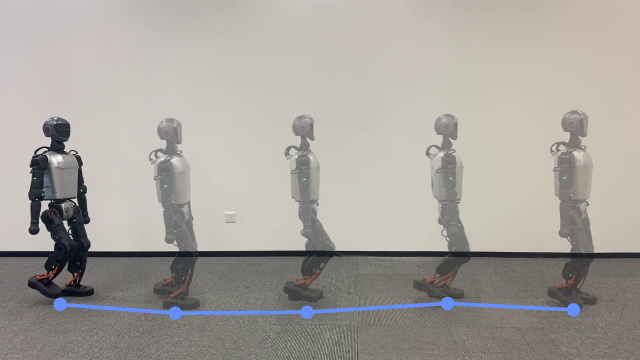} &
        \includegraphics[width=\figwidth\textwidth]{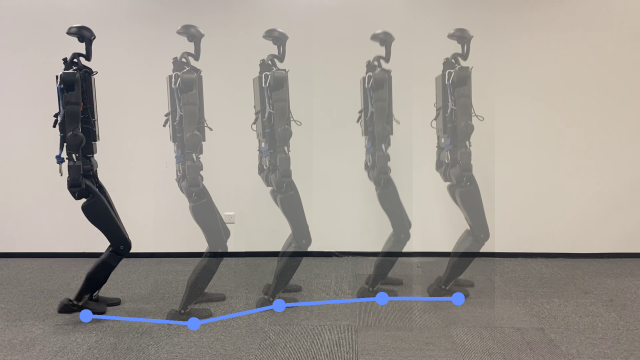} &
        \includegraphics[width=\figwidth\textwidth]{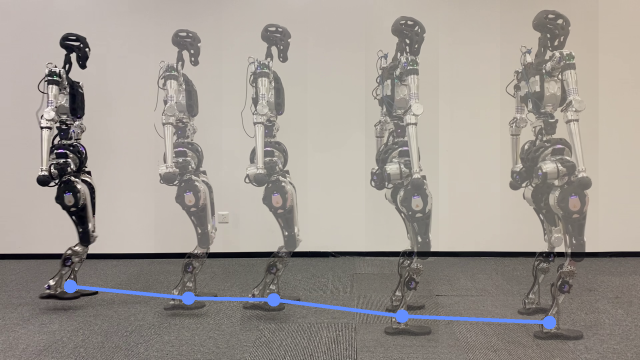} \\
        \begin{sideways}\textbf{Zero-shot}\end{sideways}
        &
        \includegraphics[width=\figwidth\textwidth]{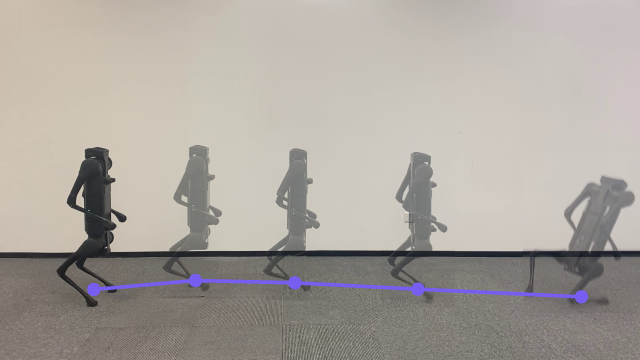} &
        \includegraphics[width=\figwidth\textwidth]{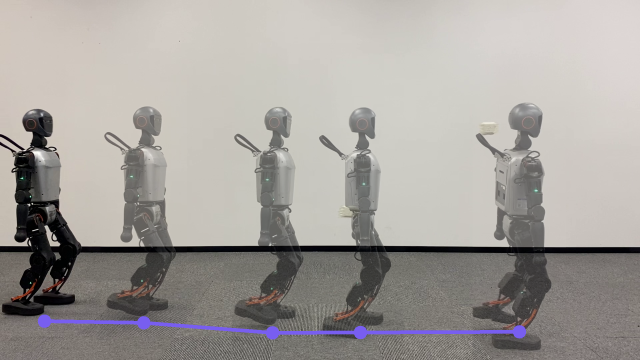} &
        \includegraphics[width=\figwidth\textwidth]{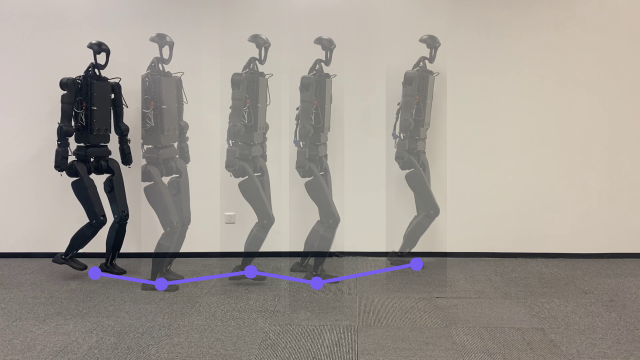} &
        \includegraphics[width=\figwidth\textwidth]{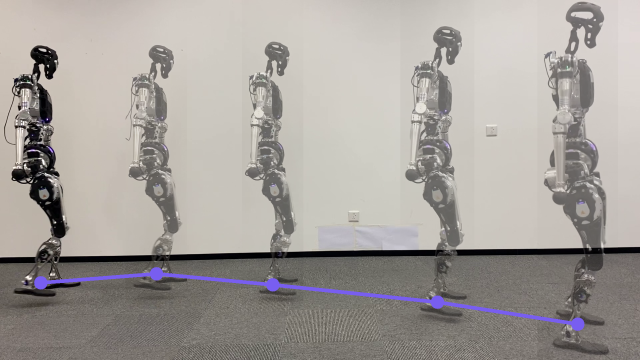} \\
        
        \begin{sideways}\textbf{Few-shot}\end{sideways}
        &
        \includegraphics[width=\figwidth\textwidth]{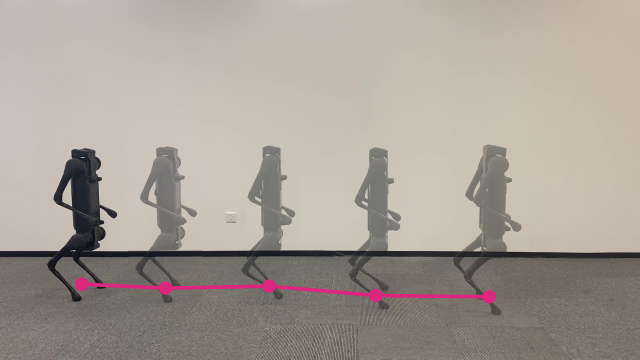} &
        \includegraphics[width=\figwidth\textwidth]{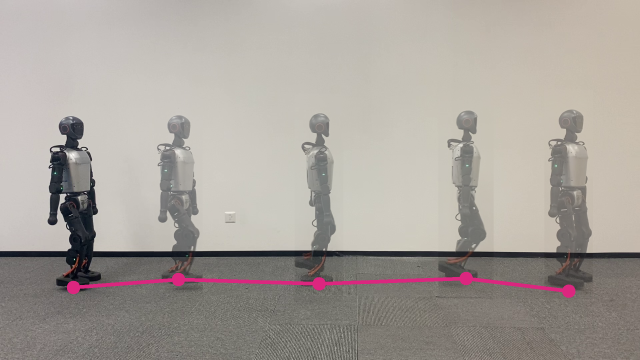} &
        \includegraphics[width=\figwidth\textwidth]{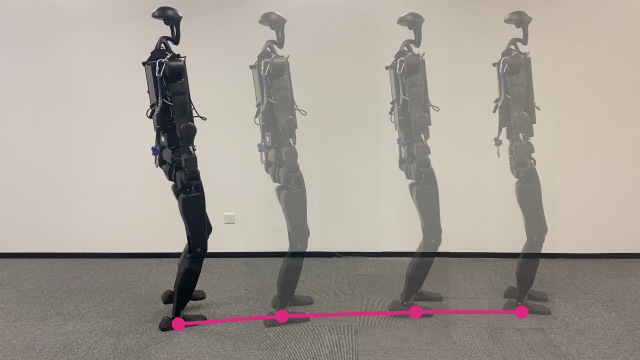} &
        \includegraphics[width=\figwidth\textwidth]{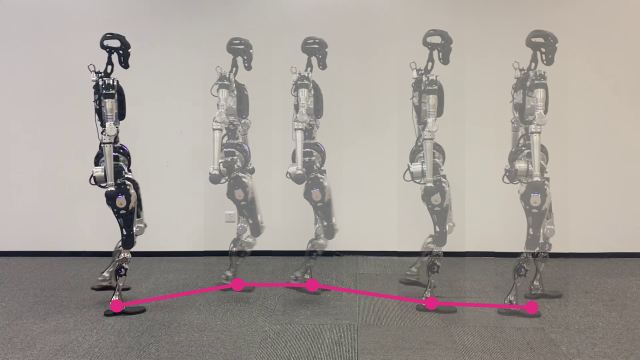} \\
        
    \end{tabular}
    \vspace{-5pt}
    \caption{
        \textbf{Snapshots from real-world deployment of learned policies.}
Columns correspond to target robots, and rows denote training regimes: scratch training (top), zero-shot transfer (middle), and few-shot adaptation (bottom). While the pretrained policy demonstrates moderate transferability to unseen robots, it often exhibits instability and velocity drift (highlighted by the line near the robot's foot). Fine-tuning mitigates these issues, resulting in more stable and consistent control.
    }
    \label{fig:real}
\end{figure*}


The results show that pretrained policies can adapt to novel embodiments within hundreds of training epochs, outperforming policies trained from scratch with thousands of epochs.
This efficiency holds across different robot morphologies, highlighting the scalability of our approach.

In \figureautorefname~\ref{fig:log}, we further validate the impact of initial sampling variance during transfer learning.
Specifically, we show that using a reduced action variance in early trajectory collection leads to much faster adaptation compared to default training settings by preserving valid locomotion behaviors from the pretrained policy.

\subsection{Real-world performance}

We evaluate the qualitative performance of the fine-tuned policies in the real world on multiple hardware platforms.
Inference runs at 50 Hz on each robot's onboard compute.
Snapshots of recorded deployments are shown in \figureautorefname~\ref{fig:real}.
The pretrained base policy demonstrates moderate adaptability to several unseen robots but is prone to jittery motion, velocity drift, and instabilities at high velocity commands.
Fine-tuning the policy for the target robot in just a few iterations enables more stable locomotion and responsive control.



\section{Conclusion}

We propose a simple yet effective pretraining method for humanoid locomotion by unifying control semantics and populating the training set with diverse embodiments that feature randomized physical parameters and action masks. This approach enables the policy to generalize across a wide range of morphologies and control interfaces.

Compared to single-robot training, the resulting pretrained policy demonstrates significantly stronger transferability to unseen robots, allowing for few-shot adaptation to novel embodiments without retraining from scratch.

Future work includes exploring curriculum-based embodiment selection and randomization scheduling to enable whole-body control.




\bibliographystyle{IEEEtran}
\bibliography{references,bibs/bib3,bibs/bib0}

@article{rma,
  title={Rma: Rapid motor adaptation for legged robots},
  author={Kumar, Ashish and Fu, Zipeng and Pathak, Deepak and Malik, Jitendra},
  journal={arXiv preprint arXiv:2107.04034},
  year={2021}
}

@inproceedings{randomization2,
  title={Dynamics randomization revisited: A case study for quadrupedal locomotion},
  author={Xie, Zhaoming and Da, Xingye and van de Panne, Michiel and Babich, Buck and Garg, Animesh},
  booktitle={2021 IEEE International Conference on Robotics and Automation (ICRA)},
  pages={4955--4961},
  year={2021},
  organization={IEEE}
}

@article{minimizing,
  title={Minimizing energy consumption leads to the emergence of gaits in legged robots},
  author={Fu, Zipeng and Kumar, Ashish and Malik, Jitendra and Pathak, Deepak},
  journal={arXiv preprint arXiv:2111.01674},
  year={2021}
}

@article{imitate,
  title={Learning agile robotic locomotion skills by imitating animals},
  author={Peng, Xue Bin and Coumans, Erwin and Zhang, Tingnan and Lee, Tsang-Wei and Tan, Jie and Levine, Sergey},
  journal={arXiv preprint arXiv:2004.00784},
  year={2020}
}

@Misc{unitree,
  title = {Unitree Robotics},
  author = {Unitree},
  howpublished = {\url{https://www.unitree.com/}},
  year = 2022,
}

@inproceedings{tert, 
address={London, United Kingdom}, 
title={Sim-to-Real Transfer for Quadrupedal Locomotion via Terrain Transformer}, 
rights={https://doi.org/10.15223/policy-029}, 
ISBN={9798350323658}, 
url={https://ieeexplore.ieee.org/document/10160497/}, 
DOI={10.1109/ICRA48891.2023.10160497}, 
booktitle={2023 IEEE International Conference on Robotics and Automation (ICRA)}, 
publisher={IEEE}, 
author={Lai, Hang and Zhang, Weinan and He, Xialin and Yu, Chen and Tian, Zheng and Yu, Yong and Wang, Jun}, 
year={2023}, 
month=may, 
pages={5141–5147}, 
language={en}
}

@article{amp-wu,
  title={Learning robust and agile legged locomotion using adversarial motion priors},
  author={Wu, Jinze and Xin, Guiyang and Qi, Chenkun and Xue, Yufei},
  journal={IEEE Robotics and Automation Letters},
  year={2023},
  publisher={IEEE}
}

@inproceedings{parkour2,
  title={Extreme parkour with legged robots},
  author={Cheng, Xuxin and Shi, Kexin and Agarwal, Ananye and Pathak, Deepak},
  booktitle={2024 IEEE International Conference on Robotics and Automation (ICRA)},
  pages={11443--11450},
  year={2024},
  organization={IEEE}
}

@inproceedings{dreamwaq,
  title={DreamWaQ: Learning robust quadrupedal locomotion with implicit terrain imagination via deep reinforcement learning},
  author={Nahrendra, I Made Aswin and Yu, Byeongho and Myung, Hyun},
  booktitle={2023 IEEE International Conference on Robotics and Automation (ICRA)},
  pages={5078--5084},
  year={2023},
  organization={IEEE}
}

@inproceedings{Daydreamer,
  title={Daydreamer: World models for physical robot learning},
  author={Wu, Philipp and Escontrela, Alejandro and Hafner, Danijar and Abbeel, Pieter and Goldberg, Ken},
  booktitle={Conference on robot learning},
  pages={2226--2240},
  year={2023},
  organization={PMLR}
}

@article{miki2022learning,
  title={Learning robust perceptive locomotion for quadrupedal robots in the wild},
  author={Miki, Takahiro and Lee, Joonho and Hwangbo, Jemin and Wellhausen, Lorenz and Koltun, Vladlen and Hutter, Marco},
  journal={Science robotics},
  volume={7},
  number={62},
  pages={eabk2822},
  year={2022},
  publisher={American Association for the Advancement of Science}
}

@article{hoeller2024anymal,
  title={Anymal parkour: Learning agile navigation for quadrupedal robots},
  author={Hoeller, David and Rudin, Nikita and Sako, Dhionis and Hutter, Marco},
  journal={Science Robotics},
  volume={9},
  number={88},
  pages={eadi7566},
  year={2024},
  publisher={American Association for the Advancement of Science}
}

@article{lee2024learning,
  title={Learning robust autonomous navigation and locomotion for wheeled-legged robots},
  author={Lee, Joonho and Bjelonic, Marko and Reske, Alexander and Wellhausen, Lorenz and Miki, Takahiro and Hutter, Marco},
  journal={Science Robotics},
  volume={9},
  number={89},
  pages={eadi9641},
  year={2024},
  publisher={American Association for the Advancement of Science}
}

@article{lee2020learning,
  title={Learning quadrupedal locomotion over challenging terrain},
  author={Lee, Joonho and Hwangbo, Jemin and Wellhausen, Lorenz and Koltun, Vladlen and Hutter, Marco},
  journal={Science robotics},
  volume={5},
  number={47},
  pages={eabc5986},
  year={2020},
  publisher={American Association for the Advancement of Science}
}

@article{gu2024advancing,
  title={Advancing humanoid locomotion: Mastering challenging terrains with denoising world model learning},
  author={Gu, Xinyang and Wang, Yen-Jen and Zhu, Xiang and Shi, Chengming and Guo, Yanjiang and Liu, Yichen and Chen, Jianyu},
  journal={arXiv preprint arXiv:2408.14472},
  year={2024}
}

@inproceedings{feng2023genloco,
  title={Genloco: Generalized locomotion controllers for quadrupedal robots},
  author={Feng, Gilbert and Zhang, Hongbo and Li, Zhongyu and Peng, Xue Bin and Basireddy, Bhuvan and Yue, Linzhu and Song, Zhitao and Yang, Lizhi and Liu, Yunhui and Sreenath, Koushil and others},
  booktitle={Conference on Robot Learning},
  pages={1893--1903},
  year={2023},
  organization={PMLR}
}

@article{he2024omnih2o,
  title={OmniH2O: Universal and Dexterous Human-to-Humanoid Whole-Body Teleoperation and Learning},
  author={He, Tairan and Luo, Zhengyi and He, Xialin and Xiao, Wenli and Zhang, Chong and Zhang, Weinan and Kitani, Kris and Liu, Changliu and Shi, Guanya},
  journal={arXiv preprint arXiv:2406.08858},
  year={2024}
}

@article{he2024hover,
  title={Hover: Versatile neural whole-body controller for humanoid robots},
  author={He, Tairan and Xiao, Wenli and Lin, Toru and Luo, Zhengyi and Xu, Zhenjia and Jiang, Zhenyu and Kautz, Jan and Liu, Changliu and Shi, Guanya and Wang, Xiaolong and others},
  journal={arXiv preprint arXiv:2410.21229},
  year={2024}
}

@article{bohlinger2024one,
  title={One policy to run them all: an end-to-end learning approach to multi-embodiment locomotion},
  author={Bohlinger, Nico and Czechmanowski, Grzegorz and Krupka, Maciej and Kicki, Piotr and Walas, Krzysztof and Peters, Jan and Tateo, Davide},
  journal={arXiv preprint arXiv:2409.06366},
  year={2024}
}

@article{huang2025learning,
  title={Learning Humanoid Standing-up Control across Diverse Postures},
  author={Huang, Tao and Ren, Junli and Wang, Huayi and Wang, Zirui and Ben, Qingwei and Wen, Muning and Chen, Xiao and Li, Jianan and Pang, Jiangmiao},
  journal={arXiv preprint arXiv:2502.08378},
  year={2025}
}

@article{wang2025beamdojo,
  title={Beamdojo: Learning agile humanoid locomotion on sparse footholds},
  author={Wang, Huayi and Wang, Zirui and Ren, Junli and Ben, Qingwei and Huang, Tao and Zhang, Weinan and Pang, Jiangmiao},
  journal={arXiv preprint arXiv:2502.10363},
  year={2025}
}

@inproceedings{huang2020,
  title={One policy to control them all: Shared modular policies for agent-agnostic control},
  author={Huang, Wenlong and Mordatch, Igor and Pathak, Deepak},
  booktitle={International Conference on Machine Learning},
  pages={4455--4464},
  year={2020},
  organization={PMLR}
}

@article{whitman2023,
  title={Learning modular robot control policies},
  author={Whitman, Julian and Travers, Matthew and Choset, Howie},
  journal={IEEE Transactions on Robotics},
  year={2023},
  publisher={IEEE}
}

@inproceedings{yu2020,
  title={Meta-world: A benchmark and evaluation for multi-task and meta reinforcement learning},
  author={Yu, Tianhe and Quillen, Deirdre and He, Zhanpeng and Julian, Ryan and Hausman, Karol and Finn, Chelsea and Levine, Sergey},
  booktitle={Conference on robot learning},
  pages={1094--1100},
  year={2020},
  organization={PMLR}
}

@article{xiao2025learning,
  title={Learning stable bipedal locomotion skills for quadrupedal robots on challenging terrains with automatic fall recovery},
  author={Xiao, Erdong and Dong, Yinzhao and Lam, James and Lu, Peng},
  journal={npj Robotics},
  volume={3},
  number={1},
  pages={22},
  year={2025},
  publisher={Nature Publishing Group UK London}
}

@Misc{fftai,
  title = {Fourier},
  author = {FourierIntelligence},
  howpublished = {\url{https://github.com/FFTAI}},
  year = 2022,
}

@Misc{g1,
  title = {G1 Humanoid Robot},
  author = {Unitree},
  howpublished = {\url{https://www.unitree.com/g1}},
  year = 2022,
}

@misc{pnd,
  title = {PNDbotics Website},
  author = {PNDbotics},
  howpublished = {\url{https://pndbotics.com/humanoid}},
  year = 2025,
}

@misc{gr1,
  title = {Fourier GR-1},
  author = {FourierIntelligence},
  howpublished = {\url{https://www.fftai.com/products-gr1}},
  year = 2025,
}

@misc{n1,
  title = {Fourier-GRX-N1 SDK},
  author = {FourierIntelligence},
  howpublished = {\url{https://fftai.github.io/fourier-grx-N1}},
  year = 2025,
}

@misc{t1,
  title = {Booster Robotics Official Website},
  author = {BoosterRobotics},
  howpublished = {\url{https://www.boosterobotics.com/robots/}},
  year = 2025,
}

@misc{pm01,
  title = {ENGINEAI-HOME},
  author = {EngineAI},
  howpublished = {\url{https://www.engineai.com.cn/product_fore}},
  year = 2025,
}

@misc{oghr2,
  title = {loongOpen/OpenLoong-Hardware},
  author = {OpenLoong},
  howpublished = {\url{https://github.com/loongOpen/OpenLoong-Hardware}},
  year = 2025,
}

@misc{kuavo,
  title = {kuavo-ros-opensource},
  author = {LejuRobotics},
  howpublished = {\url{https://gitee.com/leju-robot/kuavo-ros-opensource}},
  year = 2025,
}

@misc{atom,
  title = {DOBOT Atom},
  author = {Dobot},
  howpublished = {\url{https://www.dobot-robots.com/products/humanoid-robots/atom.html}},
  year = 2025,
}

@INPROCEEDINGS{8202133,
  author={Tobin, Josh and Fong, Rachel and Ray, Alex and Schneider, Jonas and Zaremba, Wojciech and Abbeel, Pieter},
  booktitle={2017 IEEE/RSJ International Conference on Intelligent Robots and Systems (IROS)}, 
  title={Domain randomization for transferring deep neural networks from simulation to the real world}, 
  year={2017},
  volume={},
  number={},
  pages={23-30},
  doi={10.1109/IROS.2017.8202133}}

@article{Valassakis2020CrossingTG,
  title={Crossing the Gap: A Deep Dive into Zero-Shot Sim-to-Real Transfer for Dynamics},
  author={Eugene Valassakis and Zihan Ding and Edward Johns},
  journal={2020 IEEE/RSJ International Conference on Intelligent Robots and Systems (IROS)},
  year={2020},
  pages={5372-5379},
  url={https://api.semanticscholar.org/CorpusID:221140175}
}

@inproceedings{fu2022deep,
  author    = {Fu, Zipeng and Cheng, Xuxin and Pathak, Deepak},
  title     = {Deep Whole-Body Control: Learning a Unified Policy for Manipulation and Locomotion},
  booktitle = {Conference on Robot Learning ({CoRL})},
  year      = {2022},
}

@inproceedings{he2024agile,
  author    = {He, Tairan and Zhang, Chong and Xiao, Wenli and He, Guanqi and Liu, Changliu and Shi, Guanya},
  title     = {Agile But Safe: Learning Collision-Free High-Speed Legged Locomotion},
  booktitle = {Robotics: Science and Systems (RSS)},
  year      = {2024},
}

@misc{ai2025embodimentscalinglawsrobot,
      title={Towards Embodiment Scaling Laws in Robot Locomotion}, 
      author={Bo Ai and Liu Dai and Nico Bohlinger and Dichen Li and Tongzhou Mu and Zhanxin Wu and K. Fay and Henrik I. Christensen and Jan Peters and Hao Su},
      year={2025},
      eprint={2505.05753},
      archivePrefix={arXiv},
      primaryClass={cs.RO},
      url={https://arxiv.org/abs/2505.05753}, 
}

@misc{yang2025multiloco,
      title={Multi-Loco: Unifying Multi-Embodiment Legged Locomotion via Reinforcement Learning Augmented Diffusion}, 
      author={Shunpeng Yang and Zhen Fu and Zhefeng Cao and Guo Junde and Patrick Wensing and Wei Zhang and Hua Chen},
      year={2025},
      eprint={2506.11470},
      archivePrefix={arXiv},
      primaryClass={cs.RO},
      url={https://arxiv.org/abs/2506.11470}, 
    }

@misc{yin2025unitrackerlearninguniversalwholebody,
      title={UniTracker: Learning Universal Whole-Body Motion Tracker for Humanoid Robots}, 
      author={Kangning Yin and Weishuai Zeng and Ke Fan and Zirui Wang and Qiang Zhang and Zheng Tian and Jingbo Wang and Jiangmiao Pang and Weinan Zhang},
      year={2025},
      eprint={2507.07356},
      archivePrefix={arXiv},
      primaryClass={cs.RO},
      url={https://arxiv.org/abs/2507.07356}, 
}

@article{Yu2022MultiembodimentLR,
  title={Multi-embodiment Legged Robot Control as a Sequence Modeling Problem},
  author={Chenyi Yu and Weinan Zhang and Hang Lai and Zheng Tian and Laurent Kneip and Jun Wang},
  journal={2023 IEEE International Conference on Robotics and Automation (ICRA)},
  year={2022},
  pages={7250-7257},
  url={https://api.semanticscholar.org/CorpusID:254854044}
}

@article{Patel2024GETZeroGE,
  title={GET-Zero: Graph Embodiment Transformer for Zero-shot Embodiment Generalization},
  author={Austin Patel and Shuran Song},
  journal={ArXiv},
  year={2024},
  volume={abs/2407.15002},
  url={https://api.semanticscholar.org/CorpusID:271328287}
}

@article{he2024learning,
      title={Learning human-to-humanoid real-time whole-body teleoperation},
      author={He, Tairan and Luo, Zhengyi and Xiao, Wenli and Zhang, Chong and Kitani, Kris and Liu, Changliu and Shi, Guanya},
      journal={arXiv preprint arXiv:2403.04436},
      year={2024}
    }

@article{ji2024exbody2,
  title={ExBody2: Advanced Expressive Humanoid Whole-Body Control}, 
  author={Ji, Mazeyu and Peng, Xuanbin and Liu, Fangchen and Li, Jialong and Yang, Ge and Cheng, Xuxin and Wang, Xiaolong},
  journal={arXiv preprint arXiv:2412.13196},
  year={2024},
  }

@article{cheng2024express,
title={Expressive Whole-Body Control for Humanoid Robots},
author={Cheng, Xuxin and Ji, Yandong and Chen, Junming and Yang, Ruihan and Yang, Ge and Wang, Xiaolong},
journal={arXiv preprint arXiv:2402.16796},
year={2024}
}

@article{isaacgym,
  title={Isaac Gym: High Performance GPU-Based Physics Simulation For Robot Learning},
  author={Viktor Makoviychuk and Lukasz Wawrzyniak and Yunrong Guo and Michelle Lu and Kier Storey and Miles Macklin and David Hoeller and N. Rudin and Arthur Allshire and Ankur Handa and Gavriel State},
  journal={ArXiv},
  year={2021},
  volume={abs/2108.10470}
}

@Article{	  adam,
  title		= {Adam: A Method for Stochastic Optimization},
  author	= {Diederik P. Kingma and Jimmy Ba},
  journal	= {CoRR},
  year		= {2015},
  volume	= {abs/1412.6980}
}

@Article{	  elu,
  title		= {Fast and Accurate Deep Network Learning by Exponential
		  Linear Units (ELUs)},
  author	= {Djork-Arn{\'e} Clevert and Thomas Unterthiner and Sepp
		  Hochreiter},
  journal	= {CoRR},
  year		= {2015},
  volume	= {abs/1511.07289}
}

@inproceedings{margolis2023,
  title =        {Walk these ways: Tuning robot control for generalization with
                  multiplicity of behavior},
  author =       {Margolis, Gabriel B and Agrawal, Pulkit},
  booktitle =    {Conference on Robot Learning},
  pages =        {22--31},
  year =         2023,
  organization = {PMLR}
}

@article{choi2023,
  title =        {Learning quadrupedal locomotion on deformable terrain},
  author =       {Choi, Suyoung and Ji, Gwanghyeon and Park, Jeongsoo and Kim,
                  Hyeongjun and Mun, Juhyeok and Lee, Jeong Hyun and Hwangbo,
                  Jemin},
  journal =      {Science Robotics},
  volume =       8,
  number =       74,
  pages =        {eade2256},
  year =         2023,
  publisher =    {American Association for the Advancement of Science}
}

@article{caluwaerts2023,
  title =        {Barkour: Benchmarking animal-level agility with quadruped
                  robots},
  author =       {Caluwaerts, Ken and Iscen, Atil and Kew, J Chase and Yu,
                  Wenhao and Zhang, Tingnan and Freeman, Daniel and Lee,
                  Kuang-Huei and Lee, Lisa and Saliceti, Stefano and Zhuang,
                  Vincent and others},
  journal =      {arXiv preprint arXiv:2305.14654},
  year =         2023
}

@inproceedings{peng2018,
  title =        {Sim-to-real transfer of robotic control with dynamics
                  randomization},
  author =       {Peng, Xue Bin and Andrychowicz, Marcin and Zaremba, Wojciech
                  and Abbeel, Pieter},
  booktitle =    {2018 IEEE international conference on robotics and automation
                  (ICRA)},
  pages =        {3803--3810},
  year =         2018,
  organization = {IEEE}
}

@article{campanaro2022,
  title =        {Learning and deploying robust locomotion policies with minimal
                  dynamics randomization},
  author =       {Campanaro, Luigi and Gangapurwala, Siddhant and Merkt,
                  Wolfgang and Havoutis, Ioannis},
  journal =      {arXiv preprint arXiv:2209.12878},
  year =         2022
}

@article{smith2022,
  title =        {A walk in the park: Learning to walk in 20 minutes with
                  model-free reinforcement learning},
  author =       {Smith, Laura and Kostrikov, Ilya and Levine, Sergey},
  journal =      {arXiv preprint arXiv:2208.07860},
  year =         2022
}

@article{jenelten2023,
  title =        {DTC: Deep Tracking Control--A Unifying Approach to Model-Based
                  Planning and Reinforcement-Learning for Versatile and Robust
                  Locomotion},
  author =       {Jenelten, Fabian and He, Junzhe and Farshidian, Farbod and
                  Hutter, Marco},
  journal =      {arXiv preprint arXiv:2309.15462},
  year =         2023
}

@article{bohlinger2024onepolicy,
  title =        {One Policy to Run Them All: an End-to-end Learning Approach to
                  Multi-Embodiment Locomotion},
  author =       {Bohlinger, Nico and Czechmanowski, Grzegorz and Krupka, Maciej
                  and Kicki, Piotr and Walas, Krzysztof and Peters, Jan and
                  Tateo, Davide},
  journal =      {Conference on Robot Learning},
  year =         2024
}

@inproceedings{doshi2024scaling,
  title={Scaling Cross-Embodied Learning: One Policy for Manipulation, Navigation, Locomotion and Aviation},
  author={Doshi, Ria and Walke, Homer Rich and Mees, Oier and Dasari, Sudeep and Levine, Sergey},
  booktitle={8th Annual Conference on Robot Learning},
  year={2024}
}

@article{vandermaaten08a,
  author  = {Laurens van der Maaten and Geoffrey Hinton},
  title   = {Visualizing Data using t-SNE},
  journal = {Journal of Machine Learning Research},
  year    = {2008},
  volume  = {9},
  number  = {86},
  pages   = {2579--2605},
  url     = {http://jmlr.org/papers/v9/vandermaaten08a.html}
}

@inproceedings{xue2025unified,
  title={A Unified and General Humanoid Whole-Body Controller for Fine-Grained Locomotion}, 
  author={Xue, Yufei and Dong, Wentao and Liu, Minghuan and Zhang, Weinan and Pang, Jiangmiao},
  booktitle={Robotics: Science and Systems (RSS)},
  year={2025}
}

\end{document}